# Bearing fault diagnosis based on multi-scale spectral images and convolutional neural network

Tongchao Luo, Mingquan Qiu*, Zhenyu Wu, Zebo Zhao, Dingyou Zhang

School of Physics and Mechatronic Engineering, Guizhou Minzu University, Guiyang, P.R. China.

*Corresponding author. *E-mail*: qmq_cumt@163.com

**Abstract.** To address the challenges of low diagnostic accuracy in traditional bearing fault diagnosis methods, this paper proposes a novel fault diagnosis approach based on multi-scale spectrum feature images and deep learning. Firstly, the vibration signal are preprocessed through mean removal and then converted to multi-length spectrum with fast Fourier transforms (FFT). Secondly, a novel feature called multi-scale spectral image (MSSI) is constructed by multi-length spectrum paving scheme. Finally, a deep learning framework, convolutional neural network (CNN), is formulated to diagnose the bearing faults. Two experimental cases are utilized to verify the effectiveness of the proposed method. Experimental results demonstrate that the proposed method significantly improves the accuracy of fault diagnosis.

**1. Introduction**

Rotating machinery is widely utilized in various fields such as construction, aviation, power generation, and metallurgy. Mechanical failures can result in significant economic losses and pose risks to the workers' safety[1,2]. As a critical component of rotating machinery, the reliability of rolling bearings plays a decisive role in the operational state of the machinery, particularly concerning fatigue life, friction torque, and vibration performance[3–5]. Hence it is of great importance to diagnose bearing faults immediately.

Vibration analysis is one of the most commonly used methods in the fault diagnosis of rotating machinery because the vibration signal can reflect the dynamic characteristics of the system and be both sensitive to faults and easy to measure[6]. Feature extraction and fault classification are two critical procedures in bearing fault diagnosis. Many researches have been devoted to exploring how to extract more effective features and construct more efficient identification methods for fault diagnosis in the past few decades. In existing methods, fault features are primarily constructed using time-domain, frequency-domain, and time-frequency domain feature extraction techniques, such as kurtosis analysis, cepstrum analysis, wavelet analysis, and so on [7–9]. These features typically take the form of scalars or vectors, which leads to a limited capability for health information mining. Therefore, two-dimensional (2-D) features are tentatively constructed in different ways to extract more fault information. For example, Li et al. [10] proposed to diagnose bearing faults using spectrum images of vibration signals. Guo et al. [11] extracted continuous wavelet transform scalogram as 2-D feature to conduct fault diagnosis for rotating machinery. Nie et al. [12] proposed to transform the vibration signals of multi-channel sensors into RGB images as fault features for bearing fault diagnosis. The 2-D feature based methods seem to be more robust and effective for machine fault diagnosis even in the case of limited data scarcity. Once the fault features are extracted, pattern recognition models can be employed to achieve fault diagnosis, such as statistical models, machine learning models, and so on[13,14]. In recent years, the rapid development of deep learning technologies has significantly enhanced the potential of 2-D feature based methodologies in bearing fault diagnosis. A novel research trend has emerged, which involves the conversion of vibration signals into images and the application of deep learning models for the automatic extraction of image features to facilitate fault diagnosis. This innovative approach not only effectively addresses the constraints inherent in conventional methods but also attains a superior level of diagnostic precision.

In this paper, we propose a novel fault diagnosis method using multi-scale spectral images and deep learning technology. In this method, considered that the spectral components and frequency structures of bearings typically vary with different component fault conditions, a new 2-D feature called multi-scale spectral image (MSSI) is constructed by multi-length spectrum paving strategy to depict different health states of bearings. Then a deep learning approach, convolutional neural network (CNN), is employed to model and identify faults using the extracted MSSI features. The main contributions of this work are to propose a novel feature derived directly from the spectrum with different frequency resolutions to comprehensively characterize the health state of rolling bearings, and to perform bearing fault diagnosis with deep learning technology.

The remainder of this paper is organized as follows. The process of MSSI feature extraction is explained in detail in section 2. Section 3 provides the overall procedure for bearing fault diagnosis with MSSI and CNN. Section 4 gives the experimental verification results using two bearing datasets. Finally, the conclusion of this work is given in section 5.

## 2. MSSI feature extraction

### 2.1 Overview of FFT algorithm

The Fourier transform is one of the fundamental methods for transforming time-domain series to frequency-domain. The frequency spectrum information of the signal at discrete frequency points can be obtained through DFT processing. For a discrete-time signal $x[n](n=0,1,\cdots,N-1)$, its DFT is defined as follows:

$$X[k] = \sum_{n=0}^{N-1} x[n] e^{-j\frac{2\pi}{N}nk} \tag{1}$$

where $k = 0,1,\cdots,N-1$.

Fast Fourier transform (FFT) is an efficient algorithm for computing the discrete Fourier transform (DFT). When the number of data points is $N=2^m$ (where $m$ is a positive integer), it can reduce computing complexity dramatically for analyzing data using FFT. To lay the foundation for constructing MSSI features in subsequent stages, FFT is applied for generating spectrum with different frequency resolutions, which is achieved by taking different values of $m$ in $N=2^m$ in this study.

### 2.2 Construction of the MSSI feature

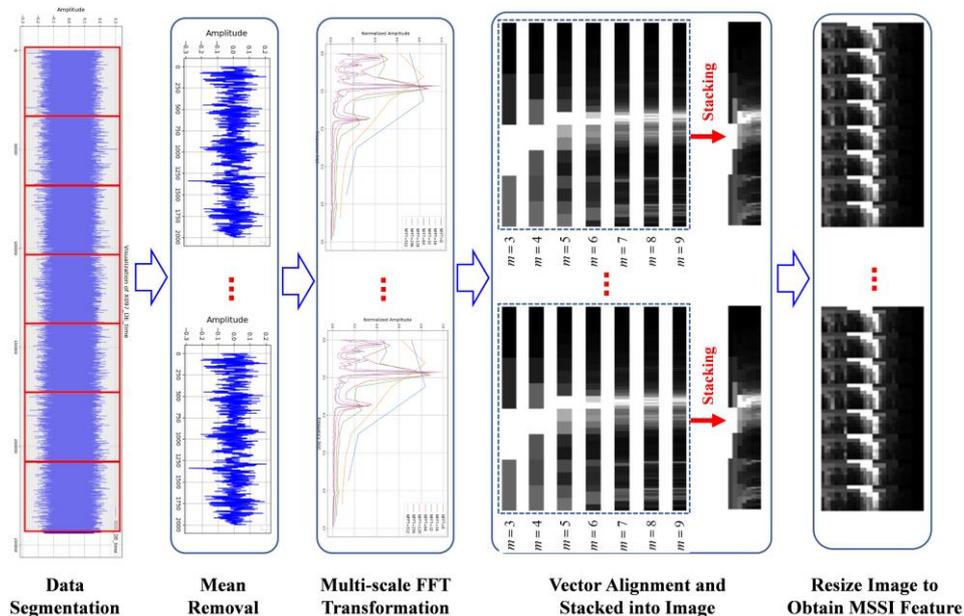

**Fig. 1.** Construction process of the proposed MSSI feature

The frequency spectrum contains a large amount of characteristic information about the running process of a bearing, such as the fault characteristic frequencies and their harmonics, the geometrical structure of the spectrum, and so on. Aiming at mining the health information contained in the spectrum as much as possible, a novel feature named MSSI is innovatively constructed by generating a series of spectrum using different frequency resolutions. The construction process of the proposed MSSI feature is presented in **Fig. 1**.

First, the collected original vibration data are first divided into sub-signals and then the signals are detrended through mean removal processing.

Second, multi-scale FFT transformation is performed to obtain the spectrum with FFT using different frequency resolutions, which achieved by setting the variable *m* in section 2.1 as different positive integers. And the variable *m* is set 3 to 9 in this paper. In addition, all spectrum sequences are pre-processed using the maximum-minimum normalization method to eliminate the effects of different magnitudes for MSSI construction.

Third, since the length of spectrum sequences vary with frequency resolutions, the vector alignment processing is performed on the aforementioned sequence. All spectrum sequences will be of the same length to facilitate subsequent construction of MSSI feature. In this study, the elements in shorter vectors are uniformly replicated to extend their lengths to match the reference sequence namely the longest one, through proportional element duplication. Then these aligned sequences are stacked into an image in order. An example of the vector alignment processing and vector stacking is presented in **Fig. 2**. Suppose that there are four series generated with $m = 4 \sim 7$, the first three sequences are aligned to the last one that served as the reference during the alignment process. Subsequently, these aligned sequences are ordered into a matrix to obtain the image feature.

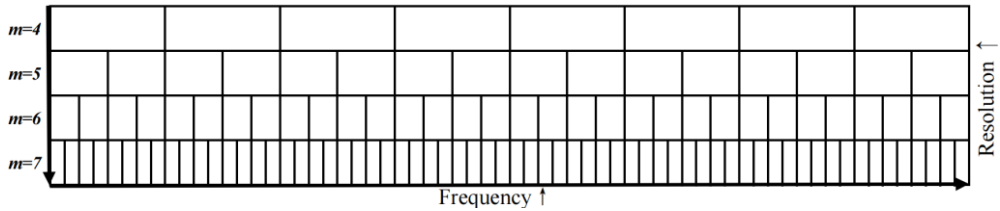

**Fig. 2.** An example of vector alignment and vectors stacking into an image

Finally, the above-mentioned image feature is usually an elongated image, which is not conducive to the further processing for the image features and the final fault diagnosis. Therefore, the image features obtained in the previous step are resized to an approximately square image (with size 32 × 56 in pixels in this study), which is called as the MSSI feature.

## 3. Fault diagnosis with MSSI and CNN

The MSSI feature, as a new 2-D feature, comprehensively depict the health information from the frequency spectrum of bearing vibration signals. Deep learning technology is employed to process these image features, and a CNN model is applied for image classification to diagnose bearing faults in our research.

### 3.1 CNN model design

This network framework is inspired by the typical LeNet-5[15] and the proposed CNN architecture is based on the same strategy, using convolutional and pooling layers to extract image features and fully connected layers for classification.

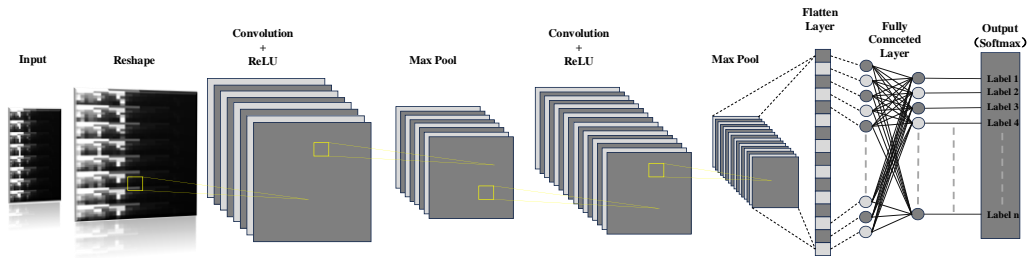

**Fig. 3.** The proposed CNN framework

Compared to the original LeNet-5, this network makes a series of changes in detail, including different number of convolutional kernels and layers, and introduces a dropout layer to mitigate the risk of overfitting[16]. To facilitate feature extraction and classification using CNN model, an input image of size 32 × 56 (in pixels) is first reshaped to a 128 × 128 (in pixels) image by nearest neighbor interpolation. The proposed CNN framework is presented in **Fig. 3**.

**Table 1.** Parameters of the proposed CNN architecture

| No. | Layer type | Number of filters | Filters size | Stride | Zero Padding | Outputs |
|---|---|---|---|---|---|---|
| 1 | Input | / | / | / | / | (128,128,1) |
| 2 | Conv1 | 16 | 3×3 | 1 | Yes | (128,128,16) |
| 3 | MaxPool | / | 2×2 | 2 | / | (64,64,16) |
| 4 | Conv2 | 32 | 3×3 | 1 | Yes | (64,64,32) |
| 5 | MaxPool | / | 2×2 | 2 | / | (32,32,32) |
| 6 | FC1 | / | / | / | / | (256,1) |
| 7 | FC2 | / | / | / | / | (10,1) |

The parameters of the proposed CNN architecture are shown in **Table 1**. In this research, we design two convolutional layers. The kernel size for both layers is 3×3, and the stride is set to 1 by default. The first convolutional layer (Conv1) uses 16 kernels, while the second convolutional layer (Conv2) contains 32 kernels, which enhances the model's feature learning capacity. Zero padding is applied to ensure that the input and output sizes remain unchanged, which adds zeros around the borders of the input volume. It is used to maintain the same spatial dimensions between the input and output volumes, thereby preserving as much information as possible from the original input[15]. The calculation formula of the convolutional layer is as follows:

$$Y(i,j) = \sum_{m=0}^{M-1} \sum_{n=0}^{N-1} X(iS+m, jS+n) \cdot W(m,n) + b \tag{2}$$

where $X(i,j)$ represents the value of the input feature at position $(i,j)$, $W(m,n)$ represents the weight of the convolution kernel at $(m,n)$ and $b$ represents the bias, $Y(i,j)$ represents the value of the output feature map at position $(i,j)$. $M \times N$ is the size of the convolution kernel and $S$ represents the stride. In convolutional layer, $S$ is set to 1 by default and $M \times N$ is set to 3×3. $P$ represents the padding and the calculation formula of the output feature map as follows:

$$H_{out} = \frac{H_{in} - M + 2P}{S} + 1 \tag{3}$$

$$W_{out} = \frac{W_{in} - N + 2P}{S} + 1 \tag{4}$$

where $H_{in}$ and $W_{in}$ represents the height and width of the input feature map, which is 128×128, $H_{out}$ and $W_{out}$ represents the height and width of the output feature map, which is 128×128, according to the formula we can get $P$ equals 1.

Additionally, two Max Pooling layers are incorporated, with a kernel size of 2×2 and a stride of 2. This pooling operation preserves the most significant features in the input feature map, while reducing its size, thereby decreasing the computational burden of subsequent layers. The expression of the maximum pooling layer in the pooling layer is as follows:

$$Y(i,j) = \max_{m,n} X(iS+m, jS+n) \tag{5}$$

where $S$ represents the stride in the pooling layer, which is set to 2, $m$ and $n$ are the length and width of the pooling kernel, which is 2×2.

After the input image passes through the convolution layer and the pooling layer, it needs to rely on the fully connected layer to classify the extracted features. In the fully connected layer, all input feature maps are expanded into one-dimensional feature vectors, weighted summed and activated by the activation function. We add two fully connected layers, FC1 and FC2, are designed, with FC1 consisting of 256 neurons and FC2 containing 10 neurons. Since FC2 serves as both the fully connected layer and the output layer, it is designed with 10 neurons to match the number of output classes. The fully connected layer can be expressed as follows[17]:

$$Y = \varphi(w^{k-1}x^{k-1} + b^k) \tag{6}$$

where $Y$ represents the output of the neuron and $\varphi(x)$ represents the activation function and $k$-1 represents the $(k-1)^{th}$ fully connected layer and $w^{k-1}$ represents the weight matrix connecting the $(k-1)^{th}$ layer to the $k^{th}$ layer and $x^{k-1}$ represents the output vector from the $(k-1)^{th}$ layer and $b$ represents the bias.

The cross-entropy loss function is applied in our research, which includes the softmax function, so the softmax is not explicitly applied in the model, and an L2 regularization parameter of 0.0001 is applied, to enhance the model performance and generalization. The cross-entropy loss function can be represented as follows[18]:

$$L = -\sum_{i=1}^{n} l_i \log(y_i) + \lambda \|W\|_2^2 \tag{7}$$

where $n$ represents the number of classes and $l_i$ represents truth label and $y_i$ represents the probability predicted by the model and $W$ represents the weight parameter matrix of the neural network and $\|W\|_2$ represents the $L2$ norm of the weight and $\lambda$ represents the weight decay, which is set to 0.0001.

A dropout layer with a dropout rate of 0.5 is introduced to prevent overfitting. The Adam optimizer is employed with a learning rate of 0.001. The hyperparameter information of the proposed CNN structure is shown in **Table 2**.

**Table 2.** Hyperparameters of the proposed CNN

| Hyperparameters | Value |
|---|---|
| Dropout | 0.5 |
| Activation Function | ReLU |
| Loss Function | Cross-Entropy Loss |
| Optimizer | Adam |
| Learning Rate | 0.001 |
| Weight Decay | 0.0001 |
| Epoch | 20 |
| Batch Size | 10 |

**3.2 The proposed fault diagnosis scheme**

This methodology includes two principal steps: offline training and online diagnosis. The diagnosis model is first established using the MSSI features derived from the offline vibration data and deep learning technology in the offline training procedure. Then the bearing health states can be diagnosed with the trained model by inputting the MSSI features derived from the online monitoring data. The main process can be summarized as follows.

(1) *MSSI Feature Extraction:* The MSSI features of the offline or the online vibration signals are constructed using the method in Section 2.

(2) *CNN Model Building:* The CNN parameters are initialized at first, and the MSSI features of the offline data are input to the CNN. Then the forward propagation algorithm is used to calculate the training error and the back propagation algorithm to obtain the error and gradients. Finally, the parameters are adjusted according to the obtained error and gradients to optimize the CNN model, thereafter a trained CNN model is obtained and can be applied for final fault diagnosis.

(2) *Online Fault Diagnosis:* The MSSI feature of an online data sample are input to the trained CNN model to diagnose the corresponding health state of bearing.

The overall flowchart of the proposed fault diagnosis scheme is shown in **Fig. 4**.

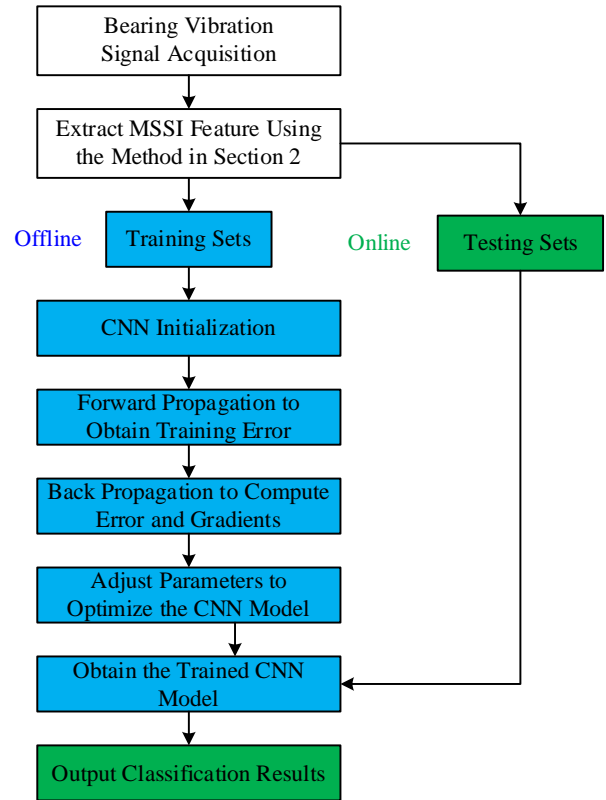

**Fig. 4.** Flow chart of fault diagnosis based on MSSI and CNN

## 4. Experimental results
### 4.1 CWRU data analysis
#### 4.1.1 Introduction to CWRU bearing data

This validation data comes from the publicly available rolling bearing dataset from Case Western Reserve University (CWRU)[19]. As shown in **Fig. 5**, the test rig consists of a 2 hp motor (left), a torque transducer/encoder (center), a dynamometer (right), and control electronics (not shown). The testing bearing were installed on the motor housing at the drive end of the motor, and the vibration signals were acquired using acceleration transducer. Four bearing health state (normal, inner race fault, rolling element fault, and outer race fault) and four levels of fault severity (7 mils, 14 mils, 21 mils and 28 mils) are considered in the experiment. And the vibration data under four different load/speed conditions (Load0 = 0 hp/1797 rpm, Load1 = 1 hp/1772 rpm, Load2 = 2 hp/1750 rpm and Load3 = 3 hp/1730 rpm) were sampled using the accelerometers attached to the drive end of the motor housing with 12 kHz sampling rate.

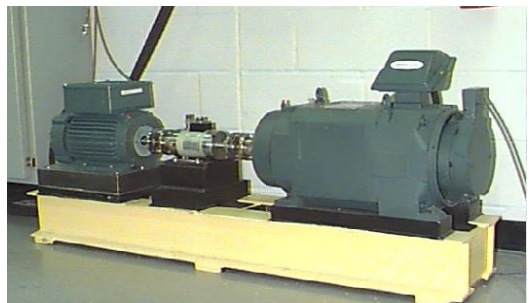

**Fig. 5.** The test rig for CWRU bearing data[19]

In order to facilitate the MSSI for extracting more fault information and enrich the data set, a data division strategy is employed for data segmentation shown in **Fig. 1**, in which a length of about 1/6 second data is applied for data segmentation. In other words, it can be guaranteed that a single sample could contain vibration data of five shaft rotations. This temporal span guarantees the comprehensive acquisition of bearing signals.

**4.1.2 Results analysis**

Here a comprehensive ten class diagnosis problem is employed to evaluate the performance of the proposed method. Several diagnosis instances under single running condition and multiple running conditions are employed for validation analysis. As shown in **Table 3**, vibration datasets under four running conditions (Load0, Load1, Load2 and Load3) are represented by A, B, C and D, respectively. And the vibration datasets under Load0–Load3 conditions is designated as E. The sample number under different running conditions is different due to the different lengths of the original data. The sample number of the normal state under Load0 is 120, while the sample number of the normal state under Load1, Load2 or Load3 is 240. The sample number of other health states is 60 or 61. Meanwhile, 70% of the total samples is randomly selected as the training set for constructing CNN model and the remaining 30% as the testing set for examining model classification ability.

**Table 3.** Description of the CWRU bearing datasets

| Health state | | | Normal | Inner race fault | | | Ball fault | | | Outer race fault | | |
|---|---|---|---|---|---|---|---|---|---|---|---|---|
| Fault diameter (mil) | | | 0 | 7 | 14 | 21 | 7 | 14 | 21 | 7 | 14 | 21 |
| Classification label | | | NM | IR1 | IR2 | IR3 | B1 | B2 | B3 | OR1 | OR2 | OR3 |
| Dataset A | Load0 | Train | 84 | 42 | 42 | 43 | 43 | 42 | 42 | 42 | 42 | 43 |
| | | Test | 36 | 18 | 18 | 18 | 18 | 18 | 18 | 18 | 18 | 18 |
| Dataset B | Load1 | Train | 168 | 42 | 42 | 42 | 42 | 43 | 42 | 43 | 43 | 42 |
| | | Test | 72 | 18 | 18 | 18 | 18 | 18 | 18 | 18 | 18 | 18 |
| Dataset C | Load2 | Train | 168 | 43 | 42 | 42 | 42 | 42 | 43 | 42 | 42 | 43 |
| | | Test | 72 | 18 | 18 | 18 | 18 | 18 | 18 | 18 | 18 | 18 |
| Dataset D | Load3 | Train | 168 | 43 | 42 | 42 | 42 | 43 | 43 | 43 | 42 | 42 |
| | | Test | 72 | 18 | 18 | 18 | 18 | 18 | 18 | 18 | 18 | 18 |
| Dataset E | Load0-Load3 | Train | 588 | 170 | 168 | 169 | 169 | 170 | 170 | 170 | 169 | 170 |
| | | Test | 252 | 72 | 72 | 72 | 72 | 72 | 72 | 72 | 72 | 72 |

The MSSI features were first extracted from both the training set and the testing set using the method presented in section 2. In order to present the proposed MSSI features intuitively, an example of MSSI feature for the 10 health states under Load1 condition is given in **Fig. 6**. It can be seen that there are relatively obvious differences between the MSSI features for different health states. The randomly selected training samples under Load0-Load3 conditions were applied to train the CNN model. An example of the training process of CNN model for Dataset E is given in **Fig. 7**. It can be seen that the loss rate drops rapidly and remains in a very small range, and both the train and validation accuracies are approximate to 100.00% after about $7^{th}$ epoch.

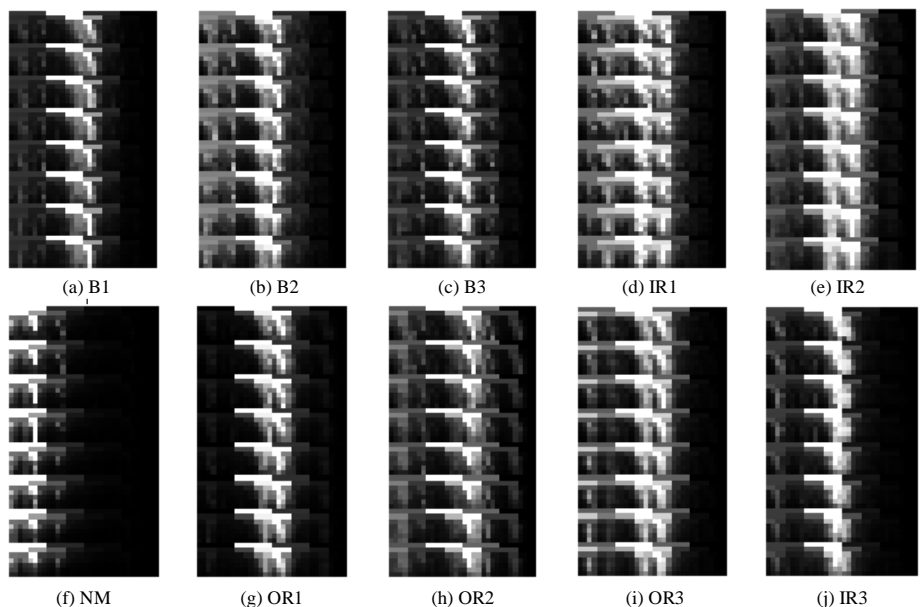

(a) B1    (b) B2    (c) B3    (d) IR1    (e) IR2

(f) NM    (g) OR1    (h) OR2    (i) OR3    (j) IR3

**Fig. 6.** An example of MSSI feature for the 10 health states under Load1 condition

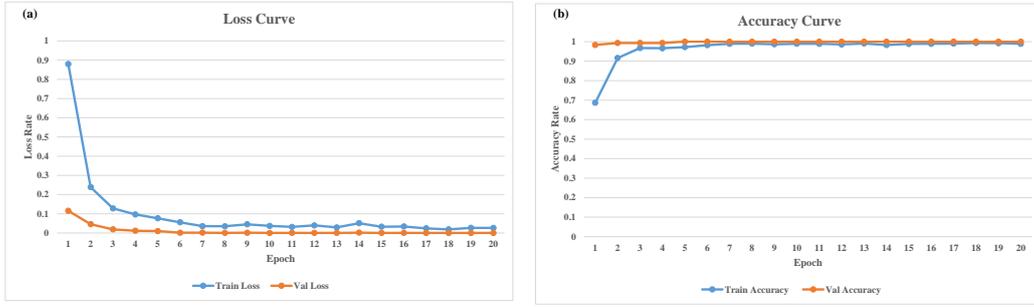

**Fig. 7.** An example of the training process of CNN model for Dataset E

Then the remaining testing samples were input to the trained CNN model to validate the effectiveness of the proposed fault diagnosis scheme. Ten trials were conducted to record the classification results and compute the average diagnostic accuracy. The accuracy results of the CWRU bearing set are presented in **Table 4** and **Table 5**. In the context of diagnostic case analysis under single running conditions, the testing samples from Dataset A, B, C and D were classified with average accuracy of 99.80%, 99.87%, 100.00% and 99.96%, respectively. For the diagnostic cases under multiple running conditions, the proposed framework exhibits robust performance with 99.99% classification accuracy for Dataset E. Examining the results, it can be observed that most testing samples are classified correctly and the proposed method demonstrates consistently high accuracy across all running conditions.

**Table 4.** Experimental results for single running condition

| Dataset | Average accuracy | Standard deviation |
|---|---|---|
| A | 99.80% | 0.350% |
| B | 99.87% | 0.400% |
| C | 100.00% | 0.000% |
| D | 99.96% | 0.140% |

**Table 5.** Experimental results for multiple running conditions

| Dataset | Testing accuracy | | | | Total accuracy | Standard deviation |
|---|---|---|---|---|---|---|
|  | Load0 | Load1 | Load2 | Load3 | | |
| E | 99.95% | 100.00% | 100.00% | 100.00% | 99.99% | 0.026% |

### 4.2 Case analysis on a conveyor idler dataset
#### 4.2.1 Introduction to the experimental platform

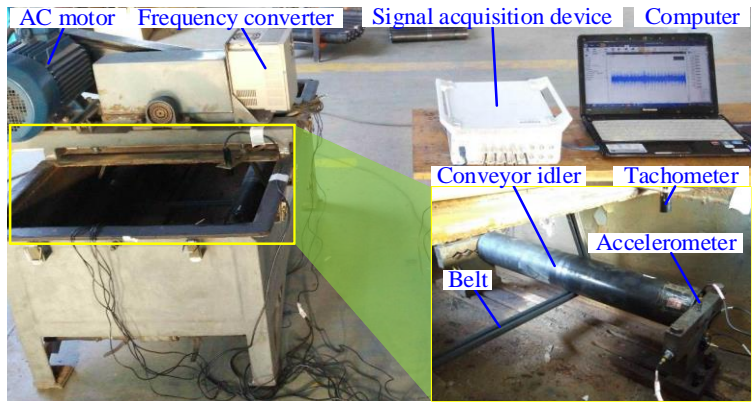

**Fig. 8.** Test bench of the belt conveyor idler[20]

The test bench shown in **Fig. 8** consists of an AC motor, a frequency converter, a conveyor idler, a tachometer, an accelerometer, a signal acquisition instrument and a computer[20]. A normal bearing (6204 type) is installed on one end and the test bearing is installed on the other end (the right end shown in **Fig. 8**) of the conveyor idler, which is driven by a belt through the AC motor. The type of both normal and tested bearings is 6204. Five bearing health states (normal, inner race fault, rolling element fault, outer race fault and cage fault) are considered in the experiment, and the vibration signals at a motor speed of 1080 rpm are collected using the sensors and the signal acquisition instrument with 20 kHz sampling frequency. One million data points (i.e., 50s) are recorded in a sampling signal and the signals are sampled twice for different bearing health states of the conveyor idler. Morea details about the experimental platform can refer to [20].

### 4.2.2 Results analysis

Similarly, to enhance the capability of MSSI in extracting fault-related information, a data division strategy is implemented for segmentation, as illustrated in **Fig. 1**. Each sample is ensured to contain vibration data corresponding to about seven shaft rotations, where this duration is sufficient to ensure the comprehensive acquisition of conveyor idler signals. Therefore, a total of 298 samples are obtained for the five health states of conveyor idler. Similarly, 70% of the total samples is randomly selected to train the CNN model and the remaining 30% is applied for testing the model diagnostic ability. The sizes of datasets for different health states are presented in **Table 6**.

Table 6. Description of the conveyor idler bearing datasets

| Label | Health state | Training set | Testing set |
|---|---|---|---|
| IR | Inner race fault | 209 | 89 |
| B | Rolling element fault | 209 | 89 |
| OR | Outer race fault | 209 | 89 |
| NM | Normal | 209 | 89 |
| CA | Cage fault | 209 | 89 |
| Total | / | 1045 | 445 |

An example of MSSI features for the conveyor idler bearing under different health states is presented in **Fig. 9.** It is evident that the characteristic differences can still be depicted between the MSSIs for different fault types. The MSSI features derived from training set were used to train the CNN model. Also, an example of the CNN training process is presented in **Fig. 10**. It can be observed that the loss rate decreases rapidly and stabilizes within a narrow range, while both the training and validation accuracies approach 100.00% after approximately the 8$^{th}$ epoch.

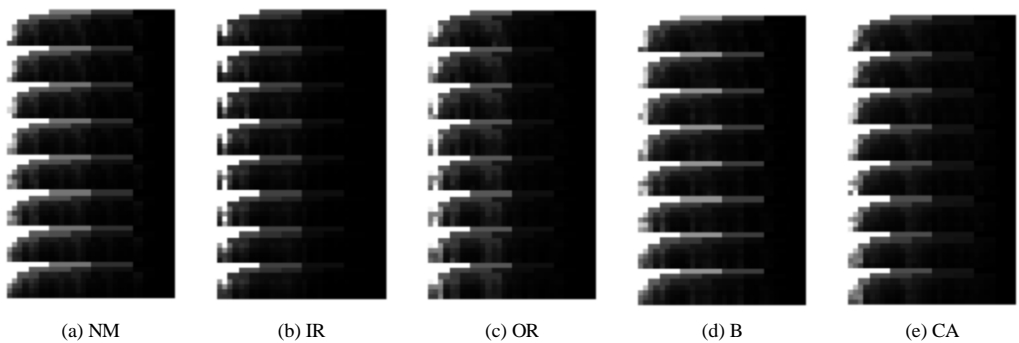

(a) NM    (b) IR    (c) OR    (d) B    (e) CA
Fig. 9. An example of MSSI features for the conveyor idler bearing under different health states

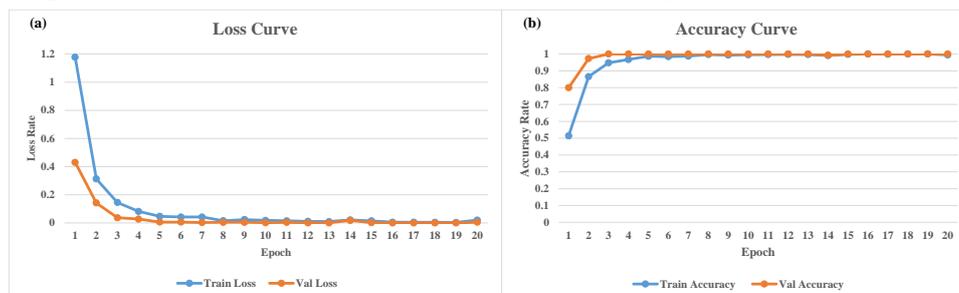

Fig. 10. An example of the training process of CNN model for conveyor idler bearing

Once the CNN model is built, the testing samples can be input to validate the performance of the proposed method. Then ten trials were conducted to obtain average results, ensuring the outcomes are generalizable and not influenced by special cases. The accuracy results are shown in **Table 7**. It can be seen that all the testing samples were classified with 100% accuracy. This further substantiates the reliability and validity of the methodology proposed in this paper.

Table 7. Accuracy results of the conveyor idler bearing analysis

| Label | IR | B | OR | NM | CA | Average accuracy | Standard deviation |
|---|---|---|---|---|---|---|---|
| Accuracy | 100.00% | 100.00% | 100.00% | 100.00% | 100.00% | 100.00% | 0.000% |

### 4.3 Discussion

In order to further compare the reliability and the potential application of the proposed scheme, a comparative study between current work and published works is presented in **Table 8**. These published methods are all verified and the ten classifications problem under single and/or multiple operational scenarios are considered with the same CWRU bearing data. As presented in **Table 8**, bearing faults were diagnosed with GAF-CA-CNN model using GAF image as feature in [21]. In [22], time-frequency images were extracted and applied for fault diagnosis, and bearing faults were diagnosed with CNN or RNELM models. In [23], ensemble deep neural network and CNN is proposed as model for bearing fault diagnosis using vibration signal and statistical feature. A comparison of the diagnosis results between this work and those published works is presented in **Table 8**.

As can be seen from **Table 8**, the proposed scheme shows better performance for bearing fault diagnosis than the others. When CNN or its related models are employed for fault classification, the proposed MSSI feature is superior to other features, even to the time-frequency image features.

Table 8. Comparison between our method and some published methods

| Reference | Method | Training | Testing | Condition | Accuracy |
|---|---|---|---|---|---|
| [21] | GAF image + GAF-CA-CNN | 9000 | 1000 | Load 0 | 99.62% |
| [22] | Time-frequency image + CNN | 4000 | 1000 | Load 0-Load 3 | 95.50% |
|  | Time-frequency image + RNELM | 4000 | 1000 | Load 0-Load 3 | 99.90% |
| [23] | Vibration signal and statistical feature + CNNPEDNN | 2000 | 370 | Load 0 | 95.76% |
|  |  | 2000 | 370 | Load 1 | 97.92% |
|  |  | 2000 | 370 | Load 2 | 97.62% |
|  |  | 2000 | 370 | Load 3 | 98.10% |
| Current work | MSSI feature + CNN | 465 | 198 | Load 0 | 99.80% |
|  |  | 549 | 234 | Load 1 | 99.87% |
|  |  | 549 | 234 | Load 2 | 100.00% |
|  |  | 550 | 234 | Load 3 | 99.96% |
|  |  | 2113 | 900 | Load 0-Load 3 | 99.99% |

### 5 Conclusion

In this paper, a novel feature, multi-scale spectral image (MSSI), is proposed as a 2-D feature to represent different health states for bearing fault diagnosis. The MSSI is constructed with FFT algorithm using multi-length spectrum paving strategy, where abundant information contained in multi-scale spectrum are deeply exploited for the representation of bearing health states. Then a new scheme is proposed for bearing fault diagnosis using a CNN architecture. The proposed method is applied to two datasets: the CWRU bearing dataset and a conveyor roller fault dataset. Compared to other fault diagnosis methods, the proposed approach demonstrates high classification performance, offering a novel solution for fault diagnosis.

**Acknowledgements**

This work was supported by the Natural Science Foundation of Guizhou Province (ZK[2021]YB270, ZK[2022]YB209), Guizhou Provincial Science and Technology Projects (ZK2024-ZD062) and the Research Foundation of Guizhou Minzu University (GZMU[2019]QN06). The authors would also like to thank Case Western Reserve University for sharing the bearing fault data on the Internet.